\documentclass[english]{gretsi}

\usepackage[utf8]{inputenc}

\usepackage[T1]{fontenc}

\usepackage{amsmath, amssymb, amsfonts}
\usepackage{amsmath,amsfonts,bm}

\def\eqref#1{equation~\ref{#1}}
\def\1{\bm{1}}

\def\vzero{{\bm{0}}}

\def\vlambda{{\bm{\lambda}}}

\def\vu{{\bm{u}}}

\def\vx{{\bm{x}}}

\def\vlambda{{\boldsymbol{\lambda}}}
\def\vbeta{{\boldsymbol{\beta}}}
\def\vrho{{\boldsymbol{\rho}}}

\def\mL{{\bm{L}}}

\def\mW{{\bm{W}}}

\DeclareMathAlphabet{\mathsfit}{\encodingdefault}{\sfdefault}{m}{sl}
\SetMathAlphabet{\mathsfit}{bold}{\encodingdefault}{\sfdefault}{bx}{n}
\newcommand{\tens}[1]{\bm{\mathsfit{#1}}}

\def\tE{{\tens{E}}}

\def\tS{{\tens{S}}}

\def\tW{{\tens{W}}}
\def\tX{{\tens{X}}}
\def\tY{{\tens{Y}}}

\def\sN{{\mathbb{N}}}

\def\sR{{\mathbb{R}}}
\def\sS{{\mathbb{S}}}

\newcommand{\R}{\mathbb{R}}

\newcommand{\limnas}{\,\xrightarrow[n\to\infty]{\text{a.s.}}\,}

\DeclareMathOperator*{\argmax}{arg\,max}

\usepackage{amsmath}
\usepackage{amssymb}
\usepackage{mathtools}
\usepackage{amsthm}
\theoremstyle{plain}
\newtheorem{theorem}{Theorem}[section]

\newtheorem{corollary}[theorem]{Corollary}
\theoremstyle{definition}

\newtheorem{assumption}[theorem]{Assumption}
\theoremstyle{remark}

\usepackage{color}
\usepackage{comment} %https://www.overleaf.com/project/640b051bcbf7438d763154ab
\usepackage{cancel}
\usepackage{url}
\usepackage{hyperref}
\usepackage{pgfplots}
\pgfplotsset{compat = newest}
\usepackage{tikz}
\usepackage{pgfplots}
\usetikzlibrary{matrix}
\usepgfplotslibrary{groupplots}
\pgfplotsset{compat=newest}
\pgfplotsset{width=7.5cm,compat=1.12}
\usepgfplotslibrary{fillbetween}
\usepackage{varioref}
\usepackage{flushend}

\usepackage[obeyFinal]{easy-todo}
\makeatletter
\def\blfootnote{\xdef\@thefnmark{}\@footnotetext}
\makeatother

\begin{document}

% ===================================================================================== %

% Titre
\titre{Hotelling Deflation on Large Symmetric Spiked Tensors}

% Auteurs
\auteurs{
  % Syntaxe : \auteur{<prénom>}{<nom>}{<adresse électronique>}{<indice d'affiliation>}
  \auteur{Mohamed El Amine}{Seddik}{mohamed.seddik@tii.ae}{1}
  \auteur{José Henrique de Morais}{Goulart}{henrique.goulart@irit.fr}{2}
  \auteur{Maxime}{Guillaud}{maxime.guillaud@inria.fr}{3}
}

% Affiliations
\affils{
  % Syntaxe : \affil{<indice d'affiliation>}{<nom du labo et adresse>}
  \affil{1}{Technology Innovation Institue,
        PO Box: 9639,
        Masdar City, Abu Dhabi, UAE
  }
  \affil{2}{IRIT, Toulouse INP, CNRS,
        2 rue Charles Camichel,
        31071 Toulouse, France
  }
  \affil{3}{Inria / CITI Laboratory,
        6 avenue des Arts,
        69621 Villeurbanne, France
  }
}

% Si tous les auteurs ont la même adresse : ne pas mettre de numéro d'affiliation. Exemple :
% \auteurs{
  % \auteur{Michel}{Dupont}{mdupont@uni.fr}{}
  % \auteur{Danielle}{Durand}{dd@uni.com}{}
% }
% \affils{
%   \affil{}{Laboratoire Traitement des Signaux,
%         1 rue de la parole, BP 00000,
%         99000 Nouvelleville Cedex 00, France
%   }
% }

% Résumé en français
\resume{Cet article étudie l'algorithme de déflation appliqué à l'estimation d'un modèle de rang faible symétrique contenu dans un tenseur de grandes dimensions corrompu par un bruit additif gaussien.
Plus précisément, nous fournissons une caractérisation précise de la performance en grandes dimensions de la déflation en termes des alignements des vecteurs obtenus par approximations successives de rang 1 et de leurs poids, en supposant des corrélations (fixes) non-triviales entre les composantes du modèle. Notre analyse permet de comprendre le mécanisme de déflation en présence de bruit et peut être exploitée pour concevoir des méthodes d'estimation plus efficaces.}

% Résumé en anglais
\abstract{This paper studies the deflation algorithm when applied to estimate a low-rank symmetric spike contained in a large tensor corrupted by additive Gaussian noise.
Specifically, we provide a precise characterization of the large-dimensional performance of deflation in terms of the alignments of the vectors obtained by successive rank-1 approximation and of their estimated weights, assuming non-trivial (fixed) correlations among spike components. Our analysis allows an understanding of the deflation mechanism in the presence of noise and can be exploited for designing more efficient signal estimation methods.} 

\maketitle

\section{ Introduction }

\blfootnote{J.~H.~de M.~Goulart's work was supported by the ANR LabEx CIMI (ANR-11-LABX-0040) within the French Programme ``Investissements d’Avenir.''.}

By capitalizing on the uniqueness properties of tensor decomposition, one can address many parameter estimation or information retrieval problems in signal processing, data sciences, and machine learning by recasting them as the decomposition of some data tensor built from the observations \cite{AnanHKT-14-JMLR}.
An archetypal application in signal processing is source separation, which can be formulated as the decomposition of a tensor containing either data acquired by a sensor array or high-order statistics estimated from these data \cite{Card-91-ICASSP}.

Among these (essentially) unique tensor decompositions, the canonical polyadic decomposition (CPD) \cite{Hitc-27-JMP} figures prominently.
It consists in writing a tensor as a (minimal) sum of rank-one terms, and as such can be seen as one extension of the singular value decomposition.
Yet, these rank-one terms need not be orthogonal for their uniqueness, and this fact is at the heart of its popularity.
In several problems, the tensor of interest is symmetric and the sought information is encoded in a rank-$r$ symmetric CPD, that is, 
\begin{equation}
 \label{eq:sym-CPD}
    \tX = \sum_{i=1}^r \beta_i \, \vx_i^{\otimes d},
\end{equation}
where $\beta_i \in \R$, $\|\vx_i\|=1$ and $\vx^{\otimes d}$ denotes the tensor product of vector $\vx \in \R^n$ with itself $d-1$ times (for instance, $\vx^{\otimes 3} = \vx \otimes \vx \otimes \vx$).
One example is in latent variable model learning \cite{AnanHKT-14-JMLR}, where the vectors $\vx_i$ appearing in the decomposition are directly related to the model's parameters (for instance, each $\vx_i$ is the mean of a Gaussian component in a mixture model).

If the vectors $\vx_i$ in (\ref{eq:sym-CPD}) were orthogonal, then one could retrieve them by resorting to a greedy deflation procedure, first introduced by Hotelling for matrices \cite{Hotelling1933}: a best rank-1 approximation of $\tX$ is computed and then subtracted from $\tX$, and the process is repeated $r$ times.
Algorithmically, each such approximation can be computed by power iteration \cite{KofiR-02-SIMAX}. 
However, this is no longer true when these vectors are not orthogonal---in fact, subtracting a best rank-1 approximation from $\tX$ can even yield a tensor of \emph{higher} rank \cite{StegeC-10-LAA}.
In some applications, this can in principle be circumvented by transforming $\tX$ in such a way that it becomes a rank-$r$ symmetric orthogonal decomposition, as long as $r \le n$.
For instance, in latent variable model learning the eigendecomposition of a matrix of second-order statistics can be exploited to obtain a whitening matrix $\mW \in \R^{r \times n}$ such that the vectors $\tilde{\vx}_i = \mW \vx_i \in \R^r$, $i=1,\ldots,r$, are pairwise orthogonal.  
An analysis of an algorithm employing this technique coupled with tensor power iteration was carried out in \cite{AnanHKT-14-JMLR}, including a robust estimation result quantifying the performance in the case one observes $\tY = \tX + \tE$, in terms of the spectral norm of the perturbation $\tE$.
Performance bounds were also derived in \cite{AnanGJ-15-COLT, AnanGJ-17-JMLR} for an algorithm involving tensor power iteration in the overcomplete regime with $r > n$, where whitening is no longer possible and thus one has to impose additional constraints to control deviation from orthogonality (such as small coherence or uniform sampling from the unit sphere). 

Yet, the results of these previous works are not well suited to the \emph{large-dimensional} regime. Specifically, the vectors $\vx_i$ become nearly orthogonal under the assumptions made in \cite{AnanGJ-15-COLT, AnanGJ-17-JMLR} as $n \to \infty$, which can be quite restrictive, while the bound on the spectral norm of the perturbation imposed on \cite{AnanHKT-14-JMLR} may not hold when the data dimension $n$ is of the same order of the number of samples used to estimate the required statistics, which is a typical assumption in this regime. 

Our goal here is to study the performance of a deflation procedure in the case where $r$ is fixed (for simplicity, $r=2$), the observations are corrupted by noise, $n \to \infty$ but the alignments $|\langle \vx_i, \vx_j \rangle| \neq 0$ are fixed and not $o(1)$.
Such an analysis is a first step towards devising more sophisticated algorithms such as orthogonalized deflation, as has been recently done in the asymmetric case \cite{SeddMD-23-arXiv}.
To this end, we build upon a recently developed approach \cite{GoulCC-22-JMLR, SeddGC-23-AAP} which allows studying random tensor models by deploying tools from random matrix theory.
The core idea of this approach is to study partial contractions, which give rise to large random matrices.

More concretely, we study the (random) alignments between the vectors obtained by deflation and the components $\vx_i$ of our CPD model, in the regime of asymptotically large tensor dimensions.
Under the assumption that these alignments (and the estimates of the weights $\beta_i$) concentrate and some additional technical conditions, we derive a system of equations that are satisfied by the limiting values of these quantities.
Once the deflation procedure is applied, one can plug its output into the equations and numerically solve for the other unknown quantities, including the weights $\beta_i$ and the alignments $\langle \vx_i, \vu_j \rangle$, where the vectors $\vu_j$ are the estimated components obtained by deflation.
Our numerical results for finite dimensions show that the obtained values closely match the predictions given by the derived equations.

\section{Spiked tensor and deflation}

\begin{table*}[t!]
    \centering
    \begin{align}\tag{S}
        \boxed{\begin{dcases*}
        \lambda_i + \frac{1}{d-1} g\left( \frac{\lambda_i}{d-1} \right) = \sum_{j=1}^r \beta_j \rho_{ij}^d - \sum_{j=1}^{i-1} \lambda_j \eta_{ij}^d, \quad
        h( \lambda_i) \rho_{ij} = \sum_{k=1}^r \beta_k \alpha_{jk} \rho_{ik}^{d-1} - \sum_{k=1}^{i-1} \lambda_k \rho_{kj} \eta_{ik}^{d-1},\quad (i,j) \in [r]^2,\\
        \left[h(\lambda_i)  + q(\lambda_j) \eta_{ij}^{d-2}\right] \eta_{ij} = \sum_{k=1}^r \beta_k \rho_{jk} \rho_{ik}^{d-1}  - \sum_{k=1}^{i-1} \lambda_k \eta_{kj} \eta_{ik}^{d-1} ,\quad  i \in [r],\, j< i.
        \end{dcases*}}
        \label{eq_system}
    \end{align}
    \vskip-4mm
\end{table*}

We consider the following rank-$r$ and order-$d$ symmetric spiked random tensor
\begin{align}
    \tS \equiv \sum_{i=1}^r \beta_i \vx_i^{\otimes d} + \frac{1}{\sqrt n} \tW,
\end{align}
with $\vx_i$ on the unit sphere $\sS^{n-1}$ and $\tW$ a $d$th-order symmetric Gaussian tensor {(see \cite{GoulCC-22-JMLR} for a formal definition)}. The signal part is modeled by the rank-$r$ component with {weights $\beta_i > 0$, which collectively determine the signal-to-noise ratio\footnote{Here we assume for simplicity that $\beta_i > 0$ for all $i \in [r]$, which implies no loss of generality for odd $d$. The case with arbitrary signs can be treated similarly, at the expense of more cumbersome derivations.} of the model}. We further assume that the rank-one components are non-orthogonal and we denote
\begin{align}
    \alpha_{ij} \equiv \langle \vx_i, \vx_j \rangle \neq 0 \quad \text{for all } i\neq j. 
\end{align}
{In the following, we will study a deflation approach aimed at approximately recovering the low-rank signal tensor, which consists} in performing successive rank-one approximations and subtracting the result at each iteration.
Specifically, at iteration $i \in [r]$, we compute the best rank-one approximation of $\tS_i$, denoted $\hat\lambda_i \vu_i^{\otimes d}$, and subtract it from $\tS_i$. Starting with $\tS_0 = \tS$, this yields the sequence of tensors
\begin{align}
\label{eq_deflation}
    \tS_{i} = \tS_{i-1} - \hat\lambda_{i-1} \vu_{i-1}^{\otimes d},
\end{align}
where {$\hat\lambda_0 = 0$ by convention, and
\begin{align}
 \label{eq:crit-pt}
    \vu_{i} \equiv & \ \argmax_{\|\vu\| = 1} \,
      \tS_{i} \cdot \vu^{d}, \\
      \hat\lambda_{i} \equiv & \ \tS_{i} \cdot \vu_i^{d},   \label{eq_lambda_hat}
\end{align}
where $\tS \cdot \vu^m$ denotes $m$-fold contraction of the tensor $\tS$ with the vector $\vu$.
% It can be shown that $(\hat\lambda_{i-1}, \vu_i)$ is a tensor eigenpair in the sense introduced in \cite{Lim-05-CAMSAP}, and thus $\tS_i \cdot \vu_i^{d-1} = \hat\lambda_i \vu_i$.
It follows that} each $\tS_i$ is also a low-rank spiked random tensor given by
\begin{align}
 \label{eq:Si-spiked}
 \textstyle 
    \tS_i = \sum_{j=1}^r \beta_j \vx_j^{\otimes d} - \sum_{j=1}^{i-1} \hat \lambda_j \vu_j^{\otimes d} + \frac{1}{\sqrt n} \tW.
\end{align}

Note that {the solution to the best rank-one tensor approximation problem (\ref{eq:crit-pt}) is in general \emph{not} a component of the CPD} of $\tS$. 
This is due to the fact that the Eckhart-Young theorem is not applicable in the non-orthogonally decomposable setting \cite{draisma2018best}. 
Thus, there is admittedly a mismatch between the objective of estimating the components of the CPD and the strategy of computing successive rank-one approximations.
Nonetheless, the deflation approach is algorithmically simple and easier to analyze than joint optimization schemes (as it relies on rank-1 approximation), and can also provide acceptable approximate solutions when cross-component correlations are small.
In the sequel, we give analytical tools to characterize and improve the accuracy achieved by Hotelling-type tensor deflation.

To understand the performance of this procedure in the large-dimensional regime, our main task consists in estimating the following quantities, which we refer to as \textit{summary statistics} as introduced by \cite{benarousSGD}, when $n \to \infty$, as functions of the parameters $\alpha_{ij}$ and $\beta_i$:
\begin{align}
 \label{eq:quant}
    \hat\lambda_i, \quad \hat \rho_{ij} \equiv \langle \vu_i, \vx_j \rangle, \quad \hat\eta_{ij} \equiv \langle \vu_i , \vu_j \rangle\quad \text{for}\,\,  i,j \in [r].
\end{align}
We will see in the sequel how this problem can be addressed through the analysis of certain random matrices, built from contractions of the tensors $\tS_i$.

\section{Main results}
{\color{darkgray}
\begin{figure*}[t!]
    \centering
    \input{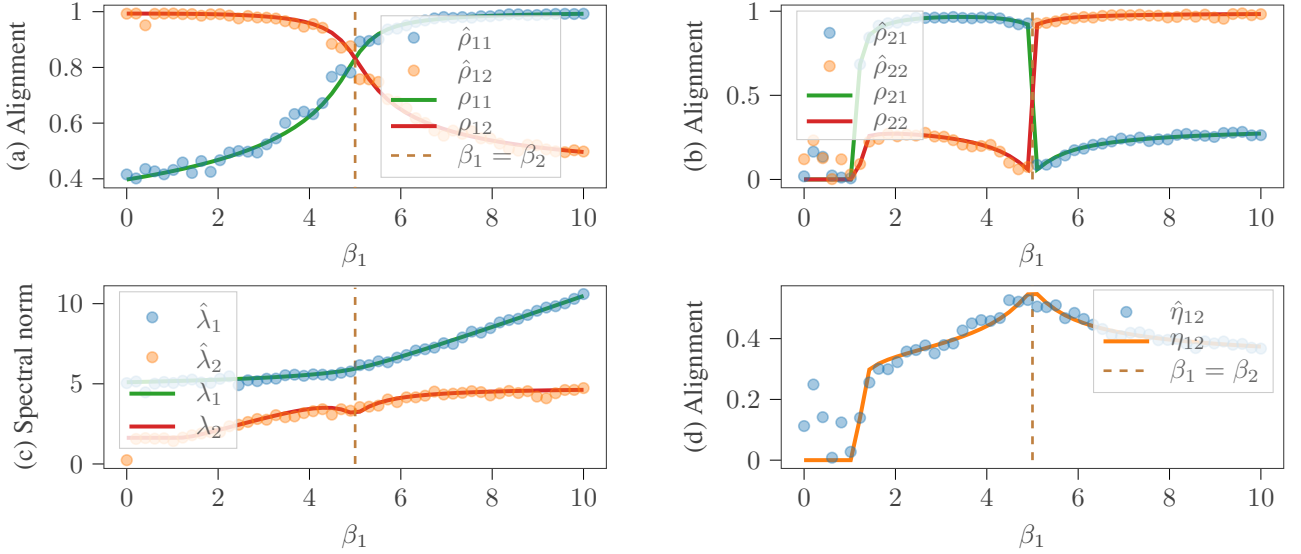}
    \caption{Empirical (dots, Monte-Carlo simulations) versus asymptotic (lines, as per Corollary \ref{cor_rank2_order3}) summary statistics of the symmetric Hotelling deflation, for parameters $r=2$, $d=3$,  $n=100$ and $\alpha=0.4$, for a range of $\beta_1$ and a fixed $\beta_2 = 5$. (a) First deflation step: alignments of $\vu_1$ with $\vx_1$ and $\vx_2$. (b) Second deflation step: alignments of $\vu_2$ with $\vx_1$ and $\vx_2$. (c) Eigenvalues $\hat \lambda_i$ and their limits $\lambda_i$ resp. (d) Alignment $\eta_{12}$ between the eigenvectors estimated at the first and second deflation step.  The asymptotic curves are obtained by solving numerically $\Psi(\cdot, \vbeta, \cdot)=\vzero$ in Corollary \ref{cor_rank2_order3} initialized with the simulated summary statistics for one realization of the noise tensor $\tW$.}
    \label{fig_alignments}
\end{figure*}
}
\subsection{Associated random matrices}

For $r=1$, the problem (\ref{eq:crit-pt}) is tantamount to the maximum likelihood estimation (MLE) of $\vx_1$. In this setting, \cite{GoulCC-22-JMLR} introduced an approach for studying the performance of MLE by borrowing tools from random matrix theory.
This approach is based upon two crucial observations: (i) critical points $\vu$ of (\ref{eq:crit-pt}) are eigenvectors of $\tS_i$ satisfying \cite{Lim-05-CAMSAP}
\begin{equation}
 \label{eq:eig}
    \tS_{i} \cdot \vu^{d-1} = \lambda \vu,
\end{equation}
with the eigenvalue $\lambda$ given by $\tS_{i} \cdot \vu^{d}$; (ii) every eigenpair $(\lambda, \vu)$ of the tensor $\tS_i$ is also an eigenpair of the matrix resulting from the contraction $\tS_{i} \cdot \vu^{d-2}$, since by (\ref{eq:eig})
\begin{equation} \label{eq_matrix_eig}
    \left(\tS_{i} \cdot \vu^{d-2}\right) \, \vu
    = \tS_{i} \cdot \vu^{d-1}
    = \lambda \vu.
\end{equation}
Hence the solution $\vu_i$ of (\ref{eq:crit-pt}) is an eigenvector (in fact, the dominant eigenvector \cite{GoulCC-22-JMLR}) of the matrix $\tS_{i} \cdot \vu_i^{d-2}$.
The matrix eigenproblem (\ref{eq_matrix_eig}) does not provide a constructive way to solve for $\vu$ since $\tS_{i} \cdot \vu^{d-2}$ itself depends on $\vu$; however, its analysis through random matrix theory allows to characterize the properties of the solutions.
Specifically, by analyzing contractions of this form, combined with the tensor eigenvalue equation (\ref{eq:eig}), \cite{GoulCC-22-JMLR} derived an asymptotic expression for the performance of MLE in terms of the alignment of $\vu_1$ and $\vx_1$ in the regime where estimation is possible (that is, beyond the phase transition characterized by \cite{JagaLM-20-AAP}).
 
Here, assuming now that $r$ is a fixed integer such that $r > 1$, we carry out a similar study of the random tensor models $\tS_i$ through the analysis of the contractions $\tS_i \cdot \vu_i^{d-2}$. 

\subsection{Limiting spectrum}
Our first result characterizes the limiting spectral measure of the contractions $\tS_i \cdot \vu_i^{d-2}$, and is instrumental in proving our main result, which is an asymptotic characterization of the summary statistics in (\ref{eq:quant}).

\begin{theorem}\label{thm_limiting_spectrum} The empirical spectral measures of $\tS_i \cdot \vu_i^{d-2}$ and of $\frac{1}{\sqrt n} \tW \cdot \vu_i^{d-2}$ converge weakly almost surely to the semi-circle distribution $\mu$ whose Stieltjes transform is given by
\begin{align*}
    g(z) \equiv \frac{2}{\gamma_d^2} \left( -z + \sqrt{z^2 - \gamma_d^2} \right),
\end{align*}
and whose density reads $\mu(dx) = \frac{2}{\pi \gamma_d^2} \sqrt{ \gamma_d^2 - x^2 } \, dx$ and is supported on $[-\gamma_d, \gamma_d]$.
\end{theorem}

\noindent\emph{Proof sketch:} The proof starts by noticing that the random matrix $\tS_i \cdot \vu_i^{d-2}$ can be written as $\mL + \frac{1}{\sqrt n} \tW \cdot \vu_i^{d-2}$ where $\mL$ is a low-rank matrix. Therefore, involving classical random matrix arguments, the matrices $\tS_i \cdot \vu_i^{d-2}$ and $\frac{1}{\sqrt n} \tW \cdot \vu_i^{d-2}$ share the same limiting spectrum, and the former is characterized similarly to the rank-one case from \cite{GoulCC-22-JMLR}.

\subsection{Limiting summary statistics}

We now look into the asymptotic values of the summary statistics introduced in (\ref{eq:quant}).
To derive them, we start from the tensor eigenvalue equations relating the pairs $(\hat\lambda_i, \vu_i)$ and the tensors $\tS_i$, that is 
{\small\begin{align}
     \hat\lambda_{i} \, \vu_i = & \ \tS_{i} \cdot \vu_i^{d-1} =  \sum_{j=1}^r \beta_j \langle \vu_i, \vx_j \rangle^{d-1} \vx_j \\
    &- \sum_{j=1}^{i-1} \hat \lambda_j \langle \vu_i, \vu_j \rangle^{d-1} \vu_j
    \nonumber  + \frac{1}{\sqrt n} \tW \cdot \vu_i^{d-1},
\end{align}}
where we used (\ref{eq:Si-spiked}).
Then, we can have access to $\hat\lambda_i$ by taking the scalar product of both sides with $\vu_i$, since this vector has a unit norm.
Similarly, $\hat\rho_{ij}$ and $\hat\eta_{ij}$ are obtained by taking scalar products with $\vx_j$ and $\vu_j$, respectively.
Next, one can compute the expectations of these quantities by invoking Stein's lemma (a.k.a.~Gaussian integration by parts) to handle the dependence between $\tW$ and each $\vu_i$, and take the limit $n \to \infty$.
Finally, similarly to \cite{GoulCC-22-JMLR, SeddGC-23-AAP, SeddMD-23-arXiv} we assume that these random quantities concentrate around their expectations, and impose some technical conditions on their limiting values, as follows.

\begin{assumption}[Almost sure convergence]\label{assump_growth} We suppose that for each tensor $\tS_i$ involved in the deflation there exists a sequence of eigenpairs $\{(\hat{\lambda}_i, \vu_i)\}_{n\in \sN}$ of $\tS_i$ such that
\begin{align*}
    \hat\lambda_i \limnas \lambda_i,\quad \hat \rho_{ij}\limnas \rho_{ij},\quad \hat \eta_{ij}\limnas \eta_{ij},
\end{align*}
with $\lambda_i > \gamma_d(d-1)$, $\rho_{ij}\neq 0$ and $\eta_{ij}\neq 0$ where $\gamma_d = 2/\sqrt{d(d-1)}$. 
\end{assumption}

Under these assumptions, we can derive a system of equations characterizing the summary statistics in the limit $n \to \infty$.
As our numerical results will show, the solutions to these equations match the empirical observations for $n$ large enough.

\begin{theorem} \label{thm_big_system_of_equations}
    Suppose that Assumption \ref{assump_growth} holds, then the limiting summary statistics $\lambda_i, \rho_{ij}$ and $\eta_{ij}$ satisfy the system of equations shown in (\ref{eq_system}) \vpageref{eq_system} with $i,j\in [r]$, where $h(z)\equiv z + g\left(z/(d-1)\right)/d$ and $q(z) \equiv g(z/(d-1)) / (d(d-1))$.
\end{theorem}

Note that Theorem \ref{thm_big_system_of_equations} only states that if the summary statistics converge to their respective limits, then the latter are solutions to the system of equations in (\ref{eq_system}); the converse is not necessarily true. In fact, studying the existence and uniqueness of the solutions of (\ref{eq_system}) is still an open question. 

As we will see in the next section, when the system (\ref{eq_system}) is solved numerically with proper initialization (e.g., with the empirical summary statistics), the obtained solutions describe well the asymptotic behavior of the \emph{maximizers} of (\ref{eq:crit-pt}), despite the fact that the tensor eigenvalue equations characterize \emph{all critical points} of these problems. We refer the reader to \cite{GoulCC-22-JMLR, SeddGC-23-AAP} for a discussion on similar phenomena observed in the rank-one case, whose rigorous explanation remains open.

\begin{corollary}\label{cor_rank2_order3}
Suppose $r=2$, $d=3$ and denote $\vlambda \equiv (\lambda_1, \lambda_2,\eta_{12})$, $\vbeta \equiv (\beta_1, \beta_2, \alpha_{12})$ and $\vrho \equiv (\rho_{ij})_{i,j\in [2]}$. Then, the limiting summary statistics $\vlambda$ and $\vrho$ satisfy $\Psi(\vlambda, \vbeta, \vrho) = \vzero$, where the mapping $\Psi:\sR^{3}\times \sR^{3} \times \sR^{4}\to \sR^7$ is defined by
\begin{small}
\begin{align*}
    \Psi \begin{pmatrix}
        \vlambda\\ \vbeta\\ \vrho
    \end{pmatrix} \equiv \begin{pmatrix}
    \sum_{i=1}^2 \beta_i \rho_{1i}^3 - f(\lambda_1)\\
    \sum_{i=1}^2 \beta_i \alpha_{1i} \rho_{1i}^2 - h(\lambda_1) \rho_{11}\\
    \sum_{i=1}^2\beta_i \alpha_{2i} \rho_{1i}^2 - h(\lambda_1) \rho_{12}\\
    \sum_{i=1}^2 \beta_i \rho_{2i}^3 - f(\lambda_2) - \lambda_1 \eta_{12}^3\\
    \sum_{i=1}^2 \beta_i \alpha_{1i} \rho_{2i}^2  - h(\lambda_2)\rho_{21} - \lambda_1 \rho_{11} \eta_{12}^2\\
    \sum_{i=1}^2 \beta_i \alpha_{2i} \rho_{2i}^2 - h(\lambda_2)\rho_{22} - \lambda_1 \rho_{12} \eta_{12}^2\\
    \sum_{i=1}^2 \beta_i \rho_{1i} \rho_{2i}^2 - h(\lambda_2)\eta_{12} - [\lambda_1 + q(\lambda_1)] \eta_{12}^2
    \end{pmatrix}.
\end{align*}
\end{small}
\end{corollary}

\section{Discussion}

Fig.~\ref{fig_alignments} illustrates the accuracy of using the deflation approach of (\ref{eq_deflation})--(\ref{eq_lambda_hat}) to estimate the spike components $\vx_i$ and weights $\beta_i$ in the correlated case $\alpha = 0.4$.
As the result depends critically on the relative values of $\beta_1$ and $\beta_2$, we let $\beta_1$ vary for a fixed $\beta_2=5$.
In Figs.~\ref{fig_alignments}(a--b), as expected from the deflation procedure, when $\beta_1<\beta_2$, $\vu_1$ tends to correlate with $\vx_2$, the strongest component, hence $\rho_{12}$ is high; conversely, for $\beta_1>\beta_2$, $\vu_1$ tends to correlate with $\vx_1$ and $\rho_{11}$ is high.
Naturally, $\rho_{21}$ and $\rho_{22}$ behave symmetrically.
Interestingly, in the regime $\beta_1 \approx \beta_2$, $\vu_1$ aligns fully neither with $\vx_1$ nor with $\vx_2$.
This indicates a significant weakness in the deflation approach with non-orthogonally decomposable tensors when several components have comparable power, since improperly estimating and subtracting the first component has the detrimental effect of \emph{increasing} the rank of the non-noise component in $\tS_{1}$ with respect to $\tS_{0}$ (see eq.~(\ref{eq:Si-spiked})).
We also note that, during the second deflation step, the estimator fails to achieve positive correlation of $\vu_2$ with either $\vx_1$ or $\vx_2$ for very low values of $\beta_1$.
Fig.~\ref{fig_alignments}(c) shows that $\lambda_1$ fairly accurately tracks the power of the strongest component (equal to $\max(\beta_1,5)$), while $\lambda_2$ is affected by a noise floor at the low range of $\beta_1$ and constitutes a poor estimator of the power of the weakest component (equal to $\min(\beta_1,5)$). 

As is common with random matrix theory, the asymptotic results from Theorem~\ref{thm_big_system_of_equations} hold approximately with remarkable accuracy for finite dimension problems thanks to the concentration of measure phenomenon\cite{benaych2011fluctuations}.

{\small\bibliography{biblio}}
\end{document}